# A Flexible Three-Dimensional Hetero-phase Computed Tomography Hepatocellular Carcinoma (HCC) Detection Algorithm for Generalizable and Practical HCC Screening

Chi-Tung Cheng[1,*], Jinzheng Cai[2,*], Wei Teng[3], Youjing Zheng[4], Yu-Ting Huang[5], Yu-Chao Wang[6], Chien-Wei Peng[3], Youbao Tang[2], Wei-Chen Lee[6], Ta-Sen Yeh[6], Jing Xiao[2], Le Lu[2], Chien-Hung Liao[1,7] and Adam P. Harrison[2]

[1] Department of Trauma and Emergency Surgery, Chang Gung Memorial Hospital, Linkou, Chang Gung University, Taoyuan, Taiwan.
[2] PAII Inc., Bethesda, MD, USA.
[3] Department of Gastroenterology and Hepatology, Chang Gung Memorial Hospital, Linkou Medical Center, Taiwan
[4] Virginia Polytechnic Institute and State University, Blacksburg, VA 24061, USA
[5] Department of Diagnostic Radiology, Chang Gung Memorial Hospital at Keelung, Chang Gung University, Taiwan.
[6] Department of General Surgery, Chang Gung Memorial Hospital, Linkou, Taiwan.
[7] Center for Artificial Intelligence in Medicine, Chang Gung Memorial Hospital, Linkou, Taoyuan, Taiwan.

* These authors contributed equally: Chi-Tung Cheng, Jinzheng Cai.
**Corresponding Authors:** Chien-Hung Liao, Department of Trauma and Emergency Surgery, Chang Gung Memorial Hospital at Linkou, 5 Fu-Shin Street, Kueishan, Taoyuan 333, Taiwan, ROC. Phone: 886-3-3281200-3651; E-mail: surgymet@gmail.com; and Adam P. Harrison, PAII Inc., 6720b Rockledge Dr, Bethesda, MD 20817. E-mail: adam.p.harrison@gmail.com.

# Abstract

Hepatocellular carcinoma (HCC) can be potentially discovered from abdominal computed tomography (CT) studies under varied clinical scenarios, e.g., fully dynamic contrast enhanced (DCE) studies, non-contrast (NC) plus venous phase (VP) abdominal studies, or NC-only studies. We develop a flexible three-dimensional deep algorithm, called hetero-phase volumetric detection (HPVD), that can accept any combination of contrast-phase inputs and with adjustable sensitivity depending on the clinical purpose.  We trained HPVD on 771 DCE CT scans to detect HCCs and tested on external 164 positives and 206 controls, respectively. We compare performance against six clinical readers, including two radiologists, two hepato-pancreatico-biliary (HPB) surgeons, and two hepatologists. The area under curve (AUC) of the localization receiver operating characteristic (LROC) for NC-only, NC plus VP, and full DCE CT yielded 0.71, 0.81, 0.89 respectively. At a high sensitivity operating point of 80% on DCE CT, HPVD achieved 97% specificity, which is comparable to measured physician performance. We also demonstrate performance improvements over more typical and less flexible non hetero-phase detectors. Thus, we demonstrate that a single deep learning algorithm can be effectively applied to diverse HCC detection clinical scenarios.

# Introduction

Hepatocellular carcinoma (HCC) is a most common primary malignant tumor of the liver and is the fourth leading cause of cancer deaths worldwide[1]. The number of new cases of HCC increases annually[1–3]. Historically, HCC was typically diagnosed late in its course because of the absence of symptoms and primary physicians' lack of alertness to provide surveillance for their high-risk patients.[3,4] HCC almost runs a fulminant course and carries a grave prognosis.

Imaging plays a key role in the diagnosis of HCC. Advances in imaging technology over the past two decades have contributed to better characterization of hepatic lesions. Dynamic contrast-enhanced (DCE) multiphase CT of the liver, which includes non-contrast (NC), arterial phase (AP), venous phase (VP), and delay phase (DP) scans, is a preferred imaging modality for surveying patients at risk of HCC[5]. The AASLD guideline recommends DCE CT as the diagnostic evaluation of HCC [6]. For large HCC lesions, reported detection sensitivity is 65%, and specificity is 96%.[7] However, when facing small lesions, sensitivity dramatically decreases to 40%. Thus, HCC lesions are at danger of being missed at an early course of the disease, when treatment is most promising. Some HCC lesions can be opportunistically found in abdominal CT studies (NC + VP) ordered for indications other than suspicious liver lesions. Even though it is less informative than DCE CT, this represents a potentially important opportunity for HCC surveillance. Most challenging of all are NC-only CT studies[5], which are necessary for patients who cannot tolerate contrast agents. Computer-aided solutions may help improve patient surveillance in these challenging settings.

Deep learning (DL) based computer-aided detection (CADe) algorithms have achieved notable successes within natural imagery.[8] In medical images, researchers have applied DL CADe to find liver lesions[9–12], but current studies either do not compare against clinical readers and/or do not enroll control patients without any target lesions; thus, gauging the CADe model screening sensitivity and specificity is difficult. Another gap is that HCC surveillance can be plausibly executed on different CT study types, each typically corresponding to its own distinct scenario. Three types are particularly prominent: DCE CT for patients at high-risk for liver lesions; NC + VP CT abdominal studies acquired for general abdominal diagnostic purposes; and NC CT studies for patients unable to tolerate contrast agents. The effectiveness of DL CADe for these different scenarios needs to be assessed. Other contrast phase combinations are also possible. Because each contrast phase can present distinct information, all acquired scans should be exploited in detection. Correspondingly, a deployed DL CADe model should be flexible enough to handle any contrast phase combination that it is presented with.

In this study, we develop a flexible DL HCC lesion CADe algorithm, called hetero-phase volumetric detection (HPVD), to deal with these challenges. Our approach integrates hetero-modal learning[13], previously used in DL segmentation models, into a powerful volumetric detection framework[11]. This hetero-modal (or hetero-phase) integration accomplishes two goals: (a) it enables HPVD to operate with any combination of CT phase inputs, providing it with maximal flexibility in deployment and (b) our single HPVD algorithm can perform better or comparably to standard input-specific CADe algorithms, each trained only on one contrast phase combination. We evaluate HPVD on DCE CT, NC + VP CT, and NC-only CT scenarios, all with a carefully selected control group, and compare its performance to six clinical readers. The HPVD algorithm could serve as a valuable means to survey for HCC in varied and distinct patient populations.

# Result

All patients were enrolled from Chang Gung Memorial Hospital (CGMH), Linkou, a major hospital in Taiwan. Figure 1 illustrates our patient and study selection process. Our target lesion type is *untreated* HCC lesions. To construct a target patient cohort, we collected 1140 DCE CT studies of patients diagnosed with an HCC lesion from histopathological analysis after either a liver resection or transplantation. We annotated each lesion using 3D bounding boxes (bboxes). Some patients could also have co-occurring HCC lesions that were treated with a chemotherapy process called transarterial chemoembolization (TACE). Such lesions are easy to identify because the TACE treated lesions retain lipiodol which shows up as hyperintense on CT images. Studies were split patient-wise into 771, 99, and 185 studies for training, validation, and testing, respectively, with no patients repeated in the test set. Patients with *only* TACE-treated HCC lesions from the pathology-proven cohort can be considered as part of the control group. To augment the control group further, we collected non-pathology-proven DCE CT studies from the same time period, selecting studies where the associated radiological report indicated no lesions. In total we selected 99 and 185 DCE CT such studies, for validation and testing respectively. DCE CT studies are principally only ordered if there is a suspected liver lesion, so these control studies sample from a challenging and representative negative population. By simply excluding the other contrast phases, the DCE CT studies also allow analysis and testing on NC + VP or NC-only settings. Table 1, Supplementary Table S1, and S2 outlines the demographic distribution of these patients.

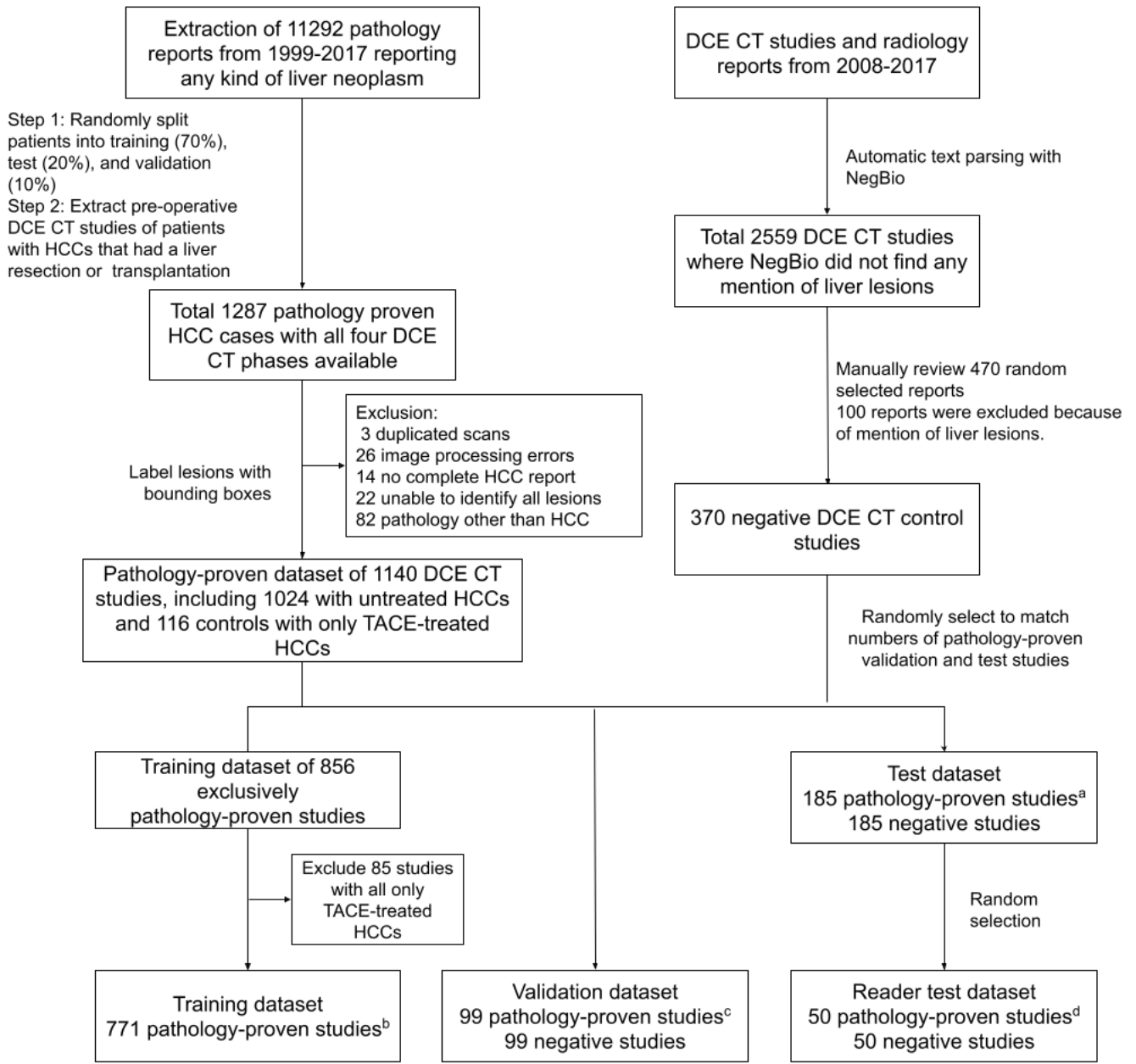

Figure. 1 Patient Selection and Dataset Preparation

Table 1. Demographic distribution of data sets. We use HCC to denote untreated HCC lesions, whereas TACE denotes TACE-treated HCC lesions. Pathology-proven and negative DCE CT studies (185 each) were enrolled as part of the *test* set. Of the pathology-proven studies, 21 were TACE-only and are considered part of the control cohort, with the remainder considered as part of the target cohort. We list characteristics of the entire test dataset first, followed by a listing of characteristics only available in the pathology-proven subset of the test set. We also list characteristics of the training set, which is entirely pathology-proven.

| **Entire Test Dataset (Pathology-Proven + Negatives)** | | | |
|---|---|---|---|
| **Feature** | **Level** | **Target** | **Control** |
| n | | 164 | 206 |
| TACE present (%) | | 7 ( 4.3) | 21 (10.2) |
| Age (std) | | 59.3 (11.1) | 53.9 (13.6) |
| Sex (%) | male | 39 (23.8) | 75 (36.4) |
| | female | 125 (76.2) | 131 (63.6) |
| **Pathology-Proven Cohorts (Target + TACE-Only Controls)** | | | |
| **Feature** | **Level** | **Training** | **Test** |
| n | | 771 | 185 |
| Total HCC lesion numbers | | 851 | 175 |
| Total TACE lesion numbers | | 73 | 50 |
| HCC makeup (%) | none (TACE-only) | 10 (1.3) | 21 (11.4) |
| | solitary | 688 (89.2) | 155 (83.8) |
| | multiple | 73 ( 9.5) | 9 ( 4.9) |
| procedure (%) | resection | 692 (89.8) | 158 (85.4) |
| | transplant | 79 (10.2) | 27 (14.6) |
| hepatitis (%) | Hep B | 417 (54.1) | 86 (46.5) |
| | Hep B + Hep C | 36 ( 4.7) | 9 ( 4.9) |
| | Hep C | 147 (19.1) | 46 (24.9) |
| | non-B non-C | 56 ( 7.3) | 17 ( 9.2) |
| | unknown | 115 (14.9) | 27 (14.6) |
| cirrhosis (%) | | 393 (51.0) | 104 (56.2) |

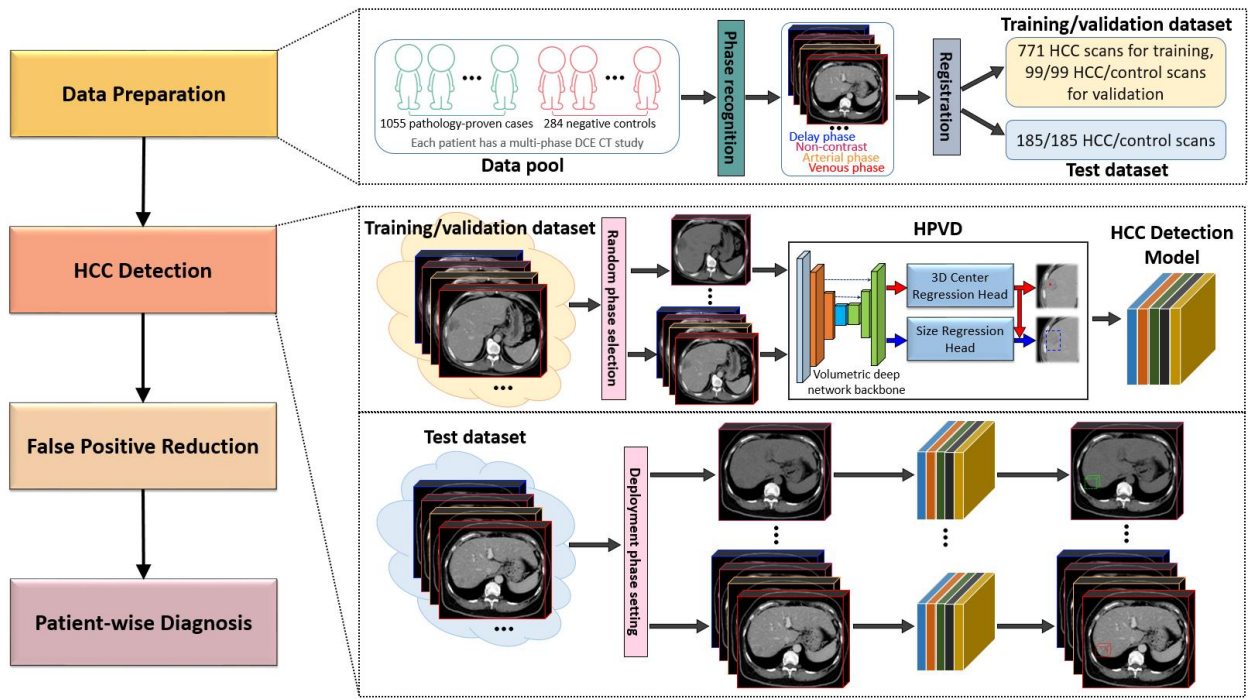

Figure. 2 The complete workflow of the image preprocessing and detection algorithm.

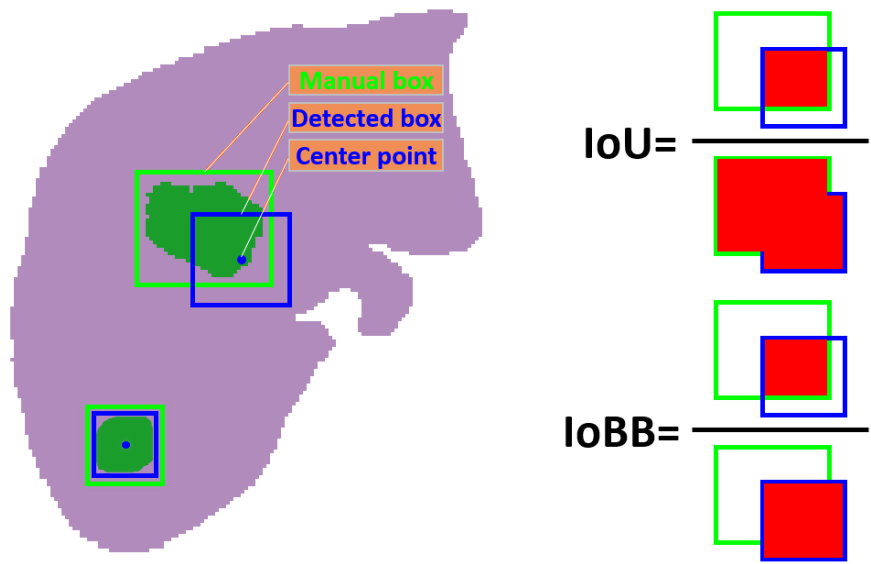

Figure 3. A bounding box detection is considered as a true positive if (a) its center falls inside the ground truth lesion bounding box; and (b) if the 3D IoBB is greater than 0.3. Here for clarity we use 2D illustrations. On the right, the overlap area used in the numerator and denominator for the IoBB metric and the more common IoU metric are pictorially illustrated.

We trained the HPVD model according to the workflow shown in Figure 2. First, we implement a preprocessing pipeline that registers the NC, AP, and DP images to the VP image. Second, we enhance and extend the state-of-the-art deep-learning VULD algorithm[11] for hetero-phase operation. VULD has been shown to outperform many popular deep-learning CADe algorithms for lesion detection[11]. Unlike VULD, a single HPVD model accepts any combination of multi-phase 3D inputs. During training, HPVD learns to predict bounding boxes (bboxes) to localize lesions from any input phase combination, which can range from single-phase to the complete four-phase DCE CT. A final post-processing identifies whether any localized lesions are untreated (target) or TACE-treated (non-target). The main goal of HPVD is to flag lesions and bring them to the attention of the physician to avoid missing observable HCCs. For evaluation, we use the criterion illustrated in Figure 3 to decide whether a predicted bbox represents a good localization or not.

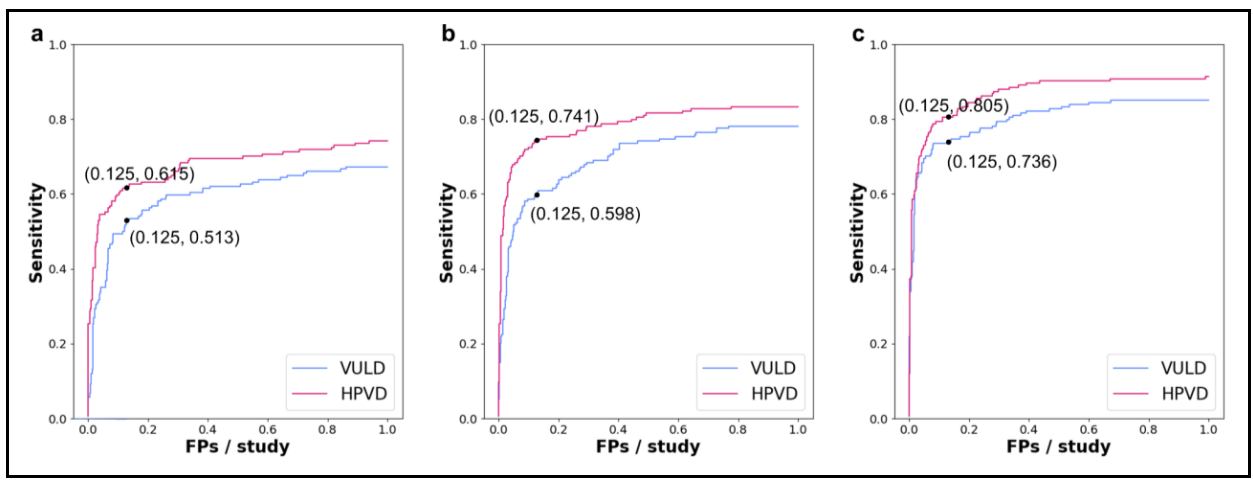

Figure 4. FROC curves of detection performance of the VULD and HPVD models. It depicts performance using the complete test set (n=370), where **a**, **b**, and **c** depict detection performance on NC-only, NC + VP, and full DCE CT studies, respectively.

Lesion-wise detection performance was evaluated using free response operating characteristic (FROC) analysis[14], which measures the relationship between detection sensitivity and the false-positives (FPs) per study. We perform FROC analysis under three different phase combinations: NC-only, NC+VP, and DCE CT, which respectively represent clinical scenarios of surveilling patients that cannot tolerate contrast, patients that may have incidental HCC findings, and patients that are at high risk of HCC. We compare a single HPVD model against three VULD models, each trained specifically for one of the three input phase combinations. Figure 4(a)-(c) renders the FROC curves for the complete test set (n=370). As can be seen, the single HPVD model outperforms each input-specific VULD model. Focusing on the 0.125 FPs/study operating point, for the NC-only, NC+VP, and DCE CT settings, HPVD achieved 61.5%, 74.1%, and 80.5% sensitivity, respectively, compared to VULD's lower sensitivities of 51.3%, 59.8%, and 73.6%, respectively. As expected, performance is best under the full DCE CT setting and sensitivities decrease as less phases are available. FROC curves for cirrhotic patients, the strongest risk factor for HCC[2], can be found in the supplementary material.

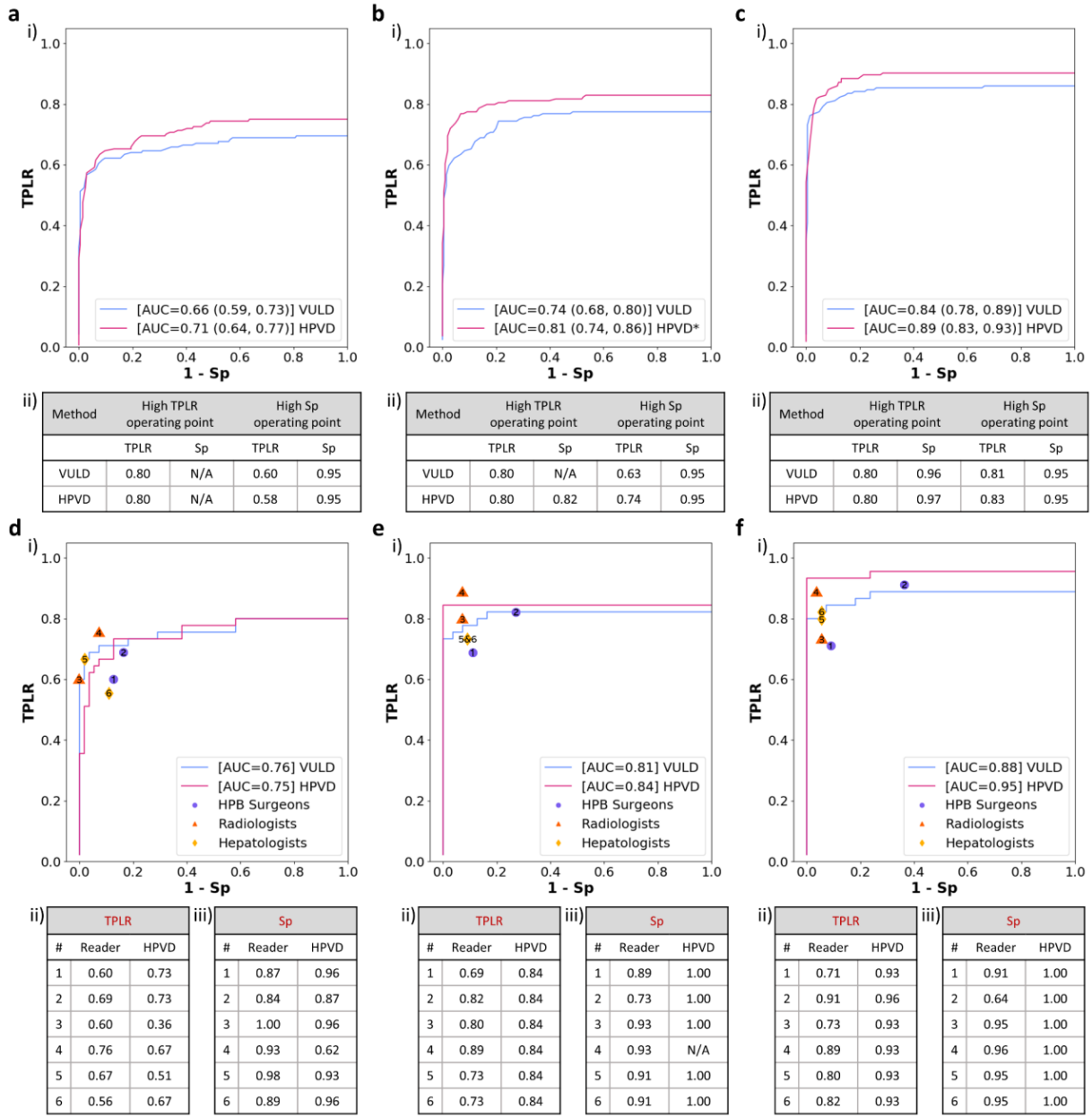

Figure 5. **Patient-wise HCC detection LROC analysis. a, b,** and **c** depict detection performance on NC-only, NC + VP, and DCE CT studies, respectively. The complete test set results (n=370) are presented, where performance of VULD and HPVD are illustrated. i, LROC curves using the top-1 detected bbox in each patient. AUCs are reported with 95% confidence intervals. The * mark in **b** indicates the differences between the VULD and HPVD AUCs are statistically significant. ii, Patient-wise LROC analysis at two possible operating points, one that could be appropriate for surveilling high-risk patients (high TPLR of 0.80) and the other for opportunistic screening of low-risk patients (high Sp

of 0.95). Some models could not achieve the high TPLR, so their Sp is set as "N/A". For **d**, **e**, and **f**, the performance is presented for the reader study subset (n=100), where reader performance is also marked. i, each data point represents a single reader, and each LROC curve represents the performance of the deep learning model using the top-1 detected bbox in each patient. ii, TPLR of each reader and the TPLR of the proposed model at a Sp chosen to match each reader's Sp. iii, Sp of each reader and the Sp of the proposed model at a TPLR chosen to match each reader's TPLR. Note, VULD requires a separate model for each input contrast phase setting. TPLR: true positive localization rate; Sp: specificity.

To evaluate patient-wise performance, we conduct localization receiver operating characteristic (LROC) analysis[15], which measures whether the maximally suspicious finding from a CADe model, for each study, localizes a true lesion or not. The LROC curve can be interpreted similarly as ROC curves, except that the ordinate of the curve is the true positive localization rate (TPLR), meaning the CADe must both (1) detect the presence of HCC lesion(s) when one or more are present and (2) localize one of them successfully. As Figure 5a-c demonstrates, when using the complete test set (n=370), the single HPVD model can yield a higher AUC than each of the three VULD models no matter the input settings. However, statistically significant improvements were only achieved for the NC + VP setting. Unsurprisingly, performance is best when full DCE CT is available, but HPVD still manages to post an 0.81 AUC when only given NC + VP CTs. The NC-only input challenges HPVD the most, but the TPLR may be high enough for opportunistic screening. Figure 5a-c also highlights the performance at high specificity and high TPLR operating points, which may be good settings for low-risk opportunistic screening and high-risk screening, respectively. As can be seen, HPVD can achieve a TPLR of 59% for the NC-only setting at a high specificity of 95%. When using full DCE CT, HPVD can reach a high TPLR of 82% at the same high specificity.

We also conducted an LROC reader study on randomly selected pathology-proven and negative cases (50 of each). Readers comprise two radiologists, two hepato-pancreatico-biliary (HPB) surgeons, and two hepatologists, who were asked to determine whether any untreated HCC lesions were present in each study and to localize the maximally suspicious instance. Figure 5d-f illustrates HPVD and reader performance for this subset. As can be seen, all readers performed well when given the four-phase DCE CT, but HPVD manages to outperform or match them. As expected, reader performance drops off substantially with the NC + VP and NC-only studies, with the two radiologists tending to perform better than the other specialties. In the NC + VP and NC-only studies, HPVD still performs well but no longer outperforms all readers.

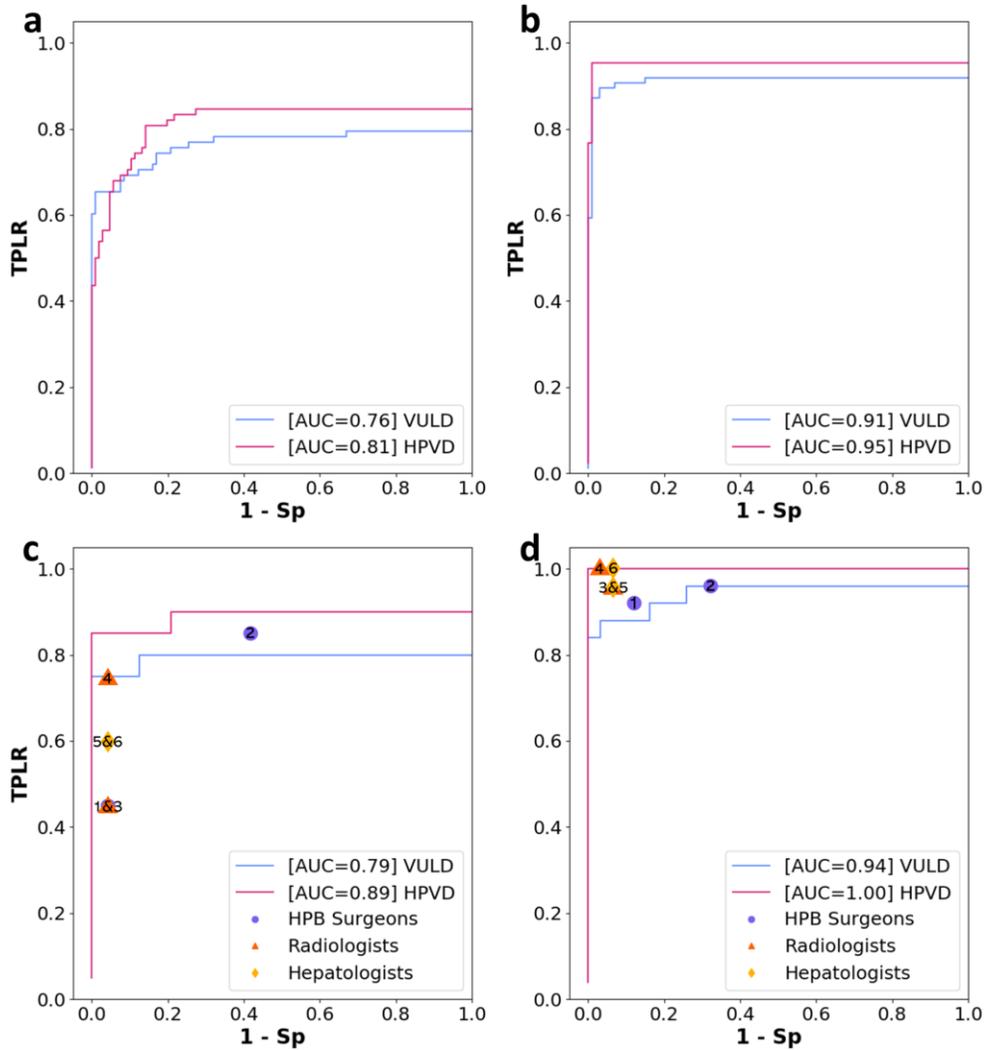

Figure 6. LROC curves using full DCE CT studies stratified by the size of the largest lesion. **a** and **b** depict results on the full test set for studies with small (n=184) and large (n=186) lesions, respectively. **c** and **d** depict results on the reader study subset for studies with small (n=44) and large (n=56) lesions, respectively, with reader performance also indicated.

Because small lesions can challenge discovery, we also evaluated the effect of lesion size on performance. To do this, we stratified the pathology-proven cohort by studies whose largest lesion is <=3cm (n=92, 49.7%) or > 3cm (n=93, 50.3%), which is a size threshold that can be used to discriminate between early and intermediate stage HCC[16]. We randomly selected negative control cases to match the number of target cases in each size stratification. As can be seen in Figure 6a-b, smaller-sized lesions indeed challenge the HPVD model, with HPVD reporting AUCs of 0.95 and 0.81 for large and small lesions, respectively, under the DCE CT setting. Figure 6c-d presents results on the reader study subset, demonstrating that such lesions also seriously challenge physician readers. Importantly, HPVD's performance is better than clinical readers, suggesting that the CADe model could be beneficial in discovering HCC lesions early in their course. The LROC curves of NC + VP and NC-only are shown in Supplementary Figure S1 and Figure S2, respectively.

Finally, Figure 7 highlights some qualitative examples. Figure 7a&b highlight cases that were localized by HPVD but missed some of the readers indicating the large interobserver variation among human readers. In Figure 7c, a HCC, sized in 1.8cm, was missed by all readers for all three input phase settings because of its small size. However, HPVD correctly localizes the lesion if it is given either abdominal CT (NC + VP) or full DCE CT. Figure 7e depicts a negative case that was correctly classified by all of the readers for all three input phase settings, whereas, HPVD incorrectly flags a part of the vessel as a HCC lesion when the DCE CT input is provided. It reveals a shortcoming of HPVD as it has not been trained to recognize vessels, however, this is relatively easy for physicians to revise such false positive findings.

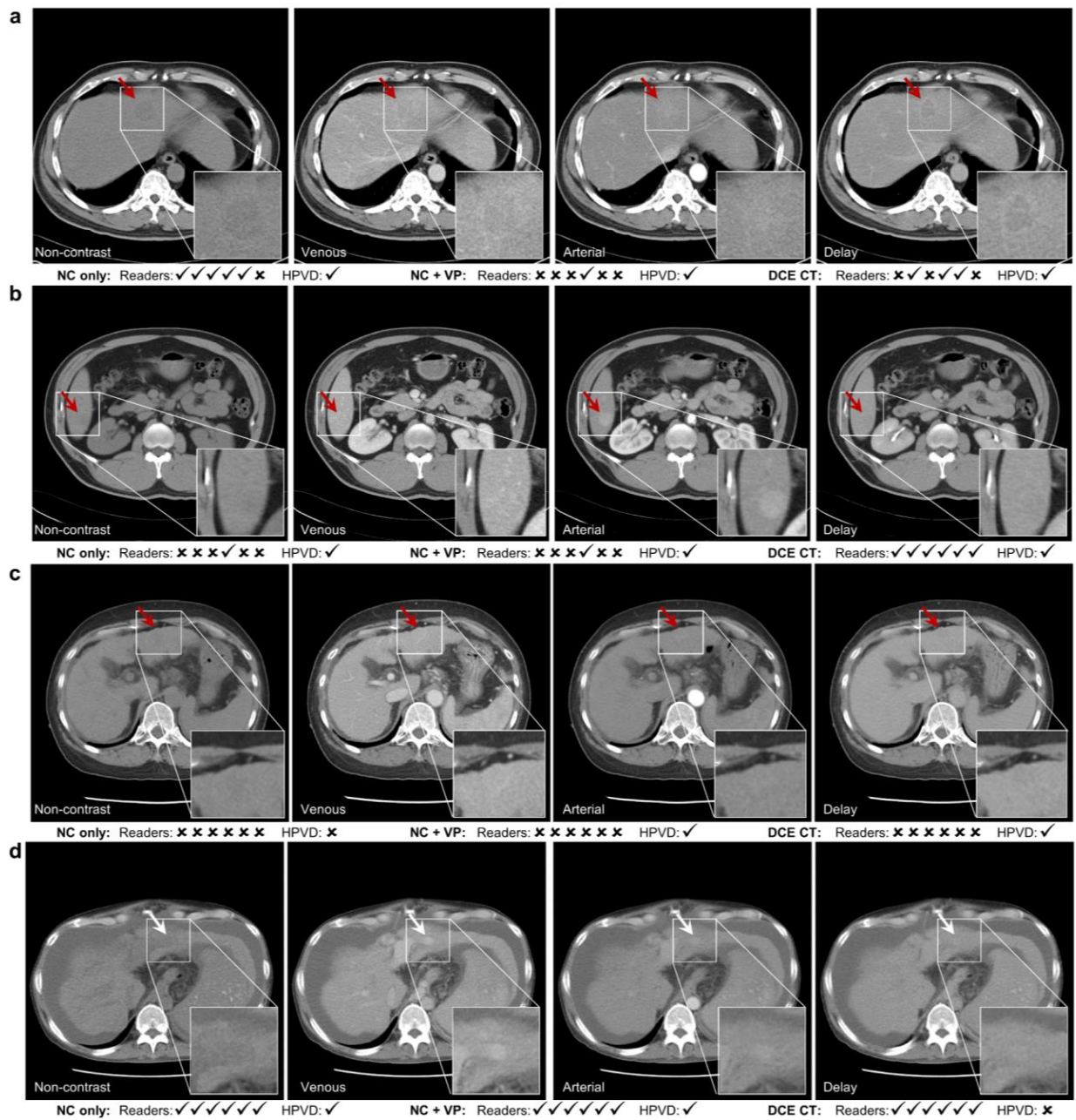

Figure 7. **Examples of HCC detection.** Each row presents DCE CT scans from one study. **a**, **b**, and **c** are studies with pathology-proven HCC and **e** is a negative study. In **a**, **b**, and **c**, the most representative axial slice of the lesion from the NC, VP, AP, and DP scans are shown from left to right. Instead of a lesion, **e** shows a false positive detection produced by HPVD when it is given DCE CT input. The patterns from each input combination indicate reader and model performance. Specifically, the two HPB surgeons, two radiologists, and two hepatologists are arranged from left to right. A check mark is used to indicate that the reader or model has correctly classified the corresponding case. Otherwise, cross marks are used to indicate incorrect classifications. Red arrows indicate HCC lesions and white arrows indicate the false positive detection in **e**.

# Discussion

This work highlights several important results. Foremost, a single HPVD CADe model can accurately localize HCC lesions in various input phase settings. When using full DCE CT images, for surveilling high-risk patients, the AUC of the LROC is 0.89. For patients scanned with standard abdominal CT (NC + VP), the AUC of LROC reaches 0.81. Even in the NC-only input setting, HPVD can yield an AUC of LROC of 0.71. On our dataset, HPVD performs comparably or better than two radiologists, two hepatologists, and two HPB surgeons, and for small lesions, HPVD performs better than all tested readers.

Effective use of CT imaging of HCC is critical for increasing patient survival, but detection is a complicated and difficult task. For one, the patient distributions involved are challenging. Unlike other liver lesions, up to 76%-90% of patients with HCC usually have cirrhosis in Western countries[19,20], causing morphological changes that can obscure lesions. Additionally, cirrhosis is associated with hepatic nodular lesions that can be easily mistaken as HCC, even by expert clinicians.[21] For cirrhotic patients, reported diagnostic sensitivity ranges from 44% to 87%.[22,23] Furthermore, the surgical or TACE treatment of HCC can cause their own obscuring morphological changes in the liver as its tissue regenerates, making it more difficult to identify viable or new tumors by CT. Our data collection protocol was carefully designed to mirror these challenges. All patients with a pathological HCC diagnosis from a resection or transplantation within 2000-2017 that had an accompanying pre-operative DCE CT study were enrolled from the CGMH, which is a major hospital and a leading center for HCC surveillance and treatment[24]. Hence, its patient population should be a good representation of the target population in East Asia, where HCC prevalence is particularly acute[16]. Our negative control population was also carefully selected, as they were drawn from patients imaged with DCE CT, which is only typically ordered for high-risk patients that match many of the profiles of the target population.

Because the liver tissue background and other anatomical structures are frequently hard to distinguish from HCC lesions, multiple contrast phases are needed to confidently detect HCC. Consequently, DCE CT is the standard diagnostic tool for HCC[6], but the use of DCE CT introduces additional difficulties. Imaging is conducted at several time intervals after contrast injection and not all images can be acquired at the right timing, which makes the contrast enhancement non-standardized across patients. The amount and rate of administration of contrast and the imaging speed also impacts DCE CT quality[17,18]. Finally, patient movement and breathing can cause non-rigid misalignments between phases. Typical identifying DCE CT features may only be present in 26-62% of HCC lesions and a non-negligible number of HCCs can be hypovascular (with poor contrast enhancement), which can lead to a false negative diagnosis[22,23].

Despite the challenges associated with DCE CT and the HCC patient population, the current HPVD algorithm can achieve a TPLR of 82% in DCE CT at a specificity of 95%, which compares well to reported clinician performance.[22,23] As Figure 5f demonstrates, HPVD can achieve a higher TPLR than all six tested readers when evaluating at their measured specificities. In terms of small lesions, and matching reports in the literature[7], reader and HPVD performance drops. However, the relative difference between HPVD and the readers (Figure 6c) is greater than for large lesions, providing further evidence of the former's clinical utility. Thus, as a support to physicians for screening or monitoring recurrence after treatment, the HPVD model may help avoid missed lesions and may help avoid additional imaging or invasive biopsy procedures for high-risk patients.

The results with DCE CT are highly promising, but DCE CT is not a frontline tool, as it takes more time to conduct and is not readily obtainable in remote centers. Thus, HCC detection applied to other CT protocols are also important to develop. For instance, standard general-purpose abdominal CT protocols consist of just the NC + VP scans. Since early-stage HCC mostly manifests without symptoms[3,4], opportunistically discovering incidental HCC findings from such protocols may be a fruitful way to detect early HCC. Solely using human clinical readers to detect incidental findings is difficult or infeasible, as the examination of the liver may not be as thorough as for CT studies ordered specifically for HCC surveillance. In such scenarios, high specificity operating points are likely critical to reduce FPs because the HCC prevalence will be lower than with high-risk populations. As Figure 5b illustrates, the HPVD TPLR is 74% at a specificity of 95% when using the NC + VP input. Apart from one radiologist, HPVD is able to outperform clinical readers for this setting (Figure 5e). HPVD could be applied to any abdominal CT study to automatically flag suspicious regions requiring more investigation. The NC-only input setting is also clinically important, as cirrhotic hepatic disease is associated with renal insufficiency[25], which can prohibit the use of contrast media to surveil HCC. Although TPLRs suffer compared to a NC + VP protocol (59% compared to 74%), HPVD can still localize HCCs at high specificities of 95%. Additionally, a high proportion of general patients who undergo medical examinations only have NC CT acquired, and incidental identification of liver tumors is not uncommon[26]. With the support of HPVD, HCC could also be detected in such scenarios.

Several prior works have proposed liver lesion CADe solutions, including some examples that have only been validated on small datasets[27,28] or did not present comparative clinical reader performance[11,12,27,28]. Zhou *et al.* did investigate reader performance against their DCE CT CADe model, but only for classification and not for localization[9]. They also only reported CADe performance at one operating point and do not specify their criteria for judging a detection as a true positive or not, which is a crucial detail for CADe evaluation[29]. Furthermore, they excluded patients with treated lesions, including TACE, and those who had surgery. Given the critical importance of detecting recurrent HCC lesions, this eliminates a major target population. Kim *et al.* reported a DCE CT CADe solution for *secondary liver metastases* from colorectal cancer and demonstrated good performance against clinical readers[10], but again no criteria were given for what constitutes a good localization. Moreover, they developed a 2D CADe model instead of a true 3D one, like ours. Because they do not have the requisite 3D context, 2D CADe models are more susceptible to FPs, which may explain why they reach a sensitivity of 81.3% at the cost of 1.3 FPs/study. By contrast, HPVD reaches 80.5% sensitivity at only 0.125 FPs/study (Figure

4c). Kim *et al.* presented a 2D CADe model and, unlike the above works, included a set of high-risk negative controls, but they did not measure against clinical readers[12]. Their 2D model exhibits a high number of ~5 FPs/study at 85% sensitivity.

In summary, none of the above works both perform a clinical reader study and include a carefully selected high-risk control cohort in their evaluation. Apart from any distinctions in evaluation, the proposed HPVD offers a notable technological distinction from all prior work: its hetero-phase capabilities. HPVD is flexible enough to handle any input phase combination setting, and we evaluate performance for NC-only, NC + VP, and DCE CT inputs. Although only three prominent input phase settings were evaluated, HPVD can, in principle, operate with any of the 15 possible phase combinations. Therefore, a single HPVD model can readily adapt to centers that do not typically acquire the DP or to studies where one or more phases happen to be missing. Standard models, like VULD[11], require a trained model for each input phase setting. Ideally, CADe models can accept whatever phases are available, with stronger performance as more inputs are given. HPVD indeed follows this trend (Figure 5a-c). Compared to three VULD models, each trained for a specific input setting, the single HPVD model sacrifices no performance. Indeed, the single HPVD model outperformed VULD in each phase setting (Figure 5a-c), although statistical significance was only achieved for the NC + VP setting. It should be emphasized that HPVD is adapted from VULD[11] and the distinction of note is the use of hetero-phase learning for the former. VULD, in turn, is a strong baseline and has been shown to outperform several leading deep learning CADe solutions[11] for 3D lesion detection. We are the first to demonstrate that a single CADe model can exhibit better performance than several input-specific models together while also being much more flexibly.

There are several limitations of this study. Our study focused specifically on detecting HCC, and not all liver lesion types, in order to focus data collection and analyses on what is currently the most prevalent malignant liver lesion. Further algorithmic development on universal hepatic lesion CADe remains an important future goal. Even if the focus remains on HCC detection, evaluating performance in the presence of other types should also be performed. Nevertheless, this presents a highly challenging data collection task as it is difficult to procure a representative enough sample of other liver lesion types, particularly since some are comparatively rare and not typically identified with gold-standard pathological diagnosis. Our target cohort also is susceptible to selection bias, as it was collected from patients that had undergone a liver resection or transplantation. Evaluations on a larger target patient cohort is key, although obtaining reliable gold-standard labels would become a challenge. Our data is currently single-center and retrospective: multi-center and prospective evaluation is another important future aim. In terms of experimental settings, our NC + VP input phase setting is meant to simulate routine abdominal CT. In reality, routine abdominal CT can use different contrast injection protocols than DCE CT, which may enhance tissues differently. Regarding the reader study, the custom CT viewing and annotation software used is not the same clinical DICOM viewer readers would have been familiar with. In addition, the scans were preprocessed with a registration algorithm that could decrease the resolution of the images. Moreover, some studies had to be re-annotated because of data transfer errors. The above factors might affect the performance of the readers and introduce a bias that differs from real-world clinical settings.

# Conclusion

In conclusion, we developed a flexible and powerful 3D HCC CADe algorithm called HPVD, which can operate with any combination of DCE CT input phases. The HPVD model demonstrated its ability by performing comparably to or better than six clinical specialists on a balanced reader test across three different input phase settings. Further prospective study is warranted to demonstrate the benefit of using HPVD for HCC surveillance in the clinical setting.

# Methods

## Patient and Image Collection

To develop and evaluate our HCC detection CADe model, we retrospectively collected two cohorts of patients: a pathology-proven cohort with HCC lesions (either untreated or already known lesions treated with transarterial chemoembolization (TACE)) and a non-pathology-proven negative cohort. These were categorized as either target (pathology-proven studies with one or more untreated HCC lesions) or control (pathology-proven studies with *only* TACE-treated lesions or negative studies). Table 1 outlines the demographic characteristics of the two datasets, whereas Figure 1 depicts a flowchart describing the patient selection process.

For the pathology-proven cohort, we examined the pathology reports indicating liver neoplasm from November 1999 to December 2017 from the CGMH, resulting in 11292 reports. From these, we identified 1287 complete reports diagnosing the presence of HCC from either hepatic resection or liver transplantation procedures that also had associated pre-operative DCE CT studies in the CGMH picture archiving and communication system (PACS) repository within 3 months prior to the procedure. The presence of co-occurring TACE-treated HCC lesions did not exclude a patient. We randomly split the dataset into training, validation, and test datasets and then manually reviewed and annotated each lesion in the studies. The pathological reports specify the number of lesions and their liver segment location. Using this information, a hepato-pancreato-biliary (HPB) surgeon (C.T.C. with 9 years experience) located every lesion in the DCE CT studies, annotating each reported lesion in each CT study with a 3D bounding box (bbox). If the reported lesion was not visible in the DCE CT scans, no bbox was drawn and the corresponding study was excluded. TACE-treated HCC lesions were also annotated, but these were recorded as a different type and not considered as an HCC annotation. After the annotation and review process, a total of 147 studies were excluded because of duplicated studies, image related problems, studies without complete HCC report, inability to identify all lesions, and the reported pathology was actually not HCC. These were split into 856, 99, and 185 studies for training, validation, and testing, respectively. Among the training and validation sets, 21 and 10 pathology-proven studies, respectively, only contained TACE-treated HCCs, resulting in 164 and 89 target studies, respectively, containing one or more untreated HCC lesions. The TACE-only studies were included as part of the control cohort. In the training dataset, we excluded studies with only TACE-treated HCC lesions for algorithmic development purposes, resulting in 771 studies. However, after a final review, 10 TACE-only studies were still found among the 771 studies.

For the negative cases, we extracted the CGMH radiological reports of DCE CT studies from 2008 to 2017, which are typically only ordered if there is a suspected liver lesion. To identify studies where no liver lesions were found, we adapted the NegBio[30] radiological report parser to flag studies where a found liver lesion was mentioned in the radiological report. Liver lesion types disqualifying a study included HCC, intrahepatic cholangiocarcinoma, hemangioma, secondary metastasis, focal nodular hyperplasia, adenoma, and abscess findings, where we used the MetaThesaurus to identify synonyms for each finding[31]. Liver cysts were not considered a positive finding. The text parser identified 2559 DCE CT studies as having no liver lesion findings, and, from these, we randomly selected 470 studies. From these, we manually confirmed that there were indeed no liver lesions findings, reducing the number to 370. We randomly selected 99 and 185 of these studies to match the numbers in the validation and test set, respectively, of the pathology-proven cohorts. The institutional review board of Chang Gung Memorial Hospital approved this study protocol (201800187B0).

## Image preprocessing

CT studies were directly extracted from the CGMH PACS, where each scan was formatted under the DICOM standard[32]. Because the scans inside each study are not curated, they can be of a very heterogeneous makeup, e.g., non-axial-reconstructed scans or non liver scans may be included[33]. Additionally, the contrast phase of each scan is also not reliably identified. As a result, acquired scans were first curated using the SE3D phase identification algorithm[33], which first visually identified all liver and axial-reconstructed scans and of those identified whether they captured a NC, AP, VP, or DP scan. To confirm the visual automatic identification, we also identified scans using simple text matching applied to the DICOM metadata. Any disagreements between SE3D and text parsing were resolved via manual inspection. Qualifying scans were then converted to the NIfTI format[34] and cropped to a common field of view.

Non-rigid misalignments between scans are common due to patient movement and breathing. These were rectified by registering the NC, AP, and DP scans to the VP scan using SimpleElastix[35] for rigid alignment followed by the DEEDS algorithm[36] for non-rigid alignment. Finally, for each study a liver mask was automatically generated using the robust multi-phase CHASe algorithm[37]. During inference, the liver mask is used to eliminate detected bboxes where less than 30% of its volume overlaps with the liver mask.

## Algorithm design

HPVD extends the volumetric universal lesion detection (VULD) deep learning architecture[11]. VULD produces volumetric detections from DCE CT scans using pseudo-3D convolutions. Specifically, VULD employs a truncated form of the 2D DenseNet-121[38] as its backbone, where the last dense block has been removed. As well, to convert the backbone for volumetric inputs, the standard 2D convolutional filters were transformed into so-called axial-coronal-saggital[39] pseudo-3D convolutions. This provides a much more computationally and memory-efficient way to process 3D CT scans compared to true 3D convolutions. Additionally, it allows using pretrained ImageNet[40] weights to initialize the backbone. To create a detector from the backbone, VULD uses the simple single-stage and anchor-free CenterNet[41] framework and loss, which has proven to be highly effective for lesion detection[11,42].

VULD handles multi-input phase scans by concatenating them together to produce a multi-channel CT. In this way, VULD can only handle one specific combination of phase inputs. Thus, to use VULD for different phase inputs, a separate model must be trained for each desired phase combination. This is inflexible, inefficient, and also deprives the model of potentially valuable variability in training. HPVD enhances VULD by adapting it for hetero-modal learning using HeMIS principles[13]. Assuming a DenseNet backbone, a standard application of HeMIS would input each DCE CT phase into initial phase-specific DenseNet blocks, $f(\cdot)$ each with their own network weights:

$$\mathbf{a}_k = f(I_k; \theta_k), (1)$$

where $I_k$ is the input CT phase, $k$ indexes the possible input phase, and $\theta_k$ are the phase-specific weights for the initial DenseNet blocks. The resulting phase-specific features, $\mathbf{a}_k$, can be collected into a set, $\mathcal{A}$. The size of the set and its composition depend on which phases are inputted. Features across phases would then be aggregated together using the mean and variance operator to create a shared feature space:

$$\mathbf{a} = f(\mathrm{mean}(\mathcal{A}); \theta_m) + f(\mathrm{var}(\mathcal{A}); \theta_v), (2)$$

where the mean and variance are computed across each phase index and the $f(\cdot; \theta_m)$ and $f(\cdot; \theta_v)$ are convolutional network layers for the mean and variance, respectively. The mean and variance operators are agnostic to the number and order of inputs, meaning any number and combination of phase features contained within $\mathcal{A}$ can be processed, which, in turn, means any combination of input CT phases. After consolidation, the combined features are then processed in a standard fashion using a single realization of the remaining DenseNet blocks.

While this is a powerful approach to accepting any combination of inputs, replicating early DenseNet network weights individually for each phase neglects a great deal of common features between contrast phases, which should be exploited. Thus, similar to the parameterization of Liu *et al.*[43], instead of making all weights in the initial DenseNet blocks phase-specific, HPVD only makes the batch normalization (BN) layers phase specific to compensate for the heterogeneity of different CT contrast phases. This includes both the BN affine weights and BN running statistics. Thus, (1) is altered to:

$$\mathbf{a}_k = f(I_k; \theta; \phi_k), ((3))$$

where $\theta$ represents the shared weights and $\phi_{k}$ represents the phase-specific BN parameters. We find this formulation to garner better performance than a standard HeMIS workflow. During training, random combinations of input phases are set as input, which equals $C(4,1) + C(4,2) + C(4,3) + C(4,4) = 15$ different possible combinations. In this way, HPVD learns how to handle any input phase combination and how to embed phase-specific features into a common feature space using the mean and variance operators.

HPVD only localizes suspicious lesion regions without discriminating their type and the patient cohorts include both untreated HCC and TACE-treated HCC lesions. Ultimately, however, we only wish to flag the former, so we must have a process to filter any detected lesions.

Fortunately, the TACE treatment causes extremely high Hounsfield units (HUs) that are not seen in untreated HCC lesions. Based on this fact, we use a simple but effective threshold-based classifier to filter out TACE-treated lesion detections: if more than 1% of CT volumes within the localized bbox have HUs greater than 200, we consider them TACE-treated. Otherwise, we consider them as untreated HCC lesions. The thresholds are picked based on our observations on the validation set. Specifically, we notice surgical clips and calcified tissue can have greater than 200 HUs, however, they usually represent less than 1% of the overall CT volume within the localized bbox of untreated HCC lesions. Finally, as is typical in CADe, we apply non-maximal suppression (NMS) to any inference results by eliminating any localized bbox if it has greater than 0.1 3D intersection over union (IoU) with another bbox of higher detection confidence.

## Lesion flagging criterion

Evaluating CADe models requires a criterion for whether a bbox represents a good localization. A common option is the IoU with the ground-truth bboxes[29]. However, a good or bad IoU does not necessarily capture whether a lesion is successfully flagged or not. For instance, a smaller predicted bbox that does not cover the entire lesion's extent may still be sufficient to successfully flag a very large lesion. IoU would unnecessarily penalize such predictions. Therefore, we use a different criterion, which we call the "lesion flagging criterion." First, the center of a predicted bbox must lie inside the ground-truth bbox, which is sometimes called the pointing game in machine learning literature[45]. Second, a predicted bbox must also have a high enough intersection over the detected bbox area (IoBB)[46], which does not penalize predicted bboxes smaller in area than the ground truth bbox. We use a 3D IoBB threshold of >= 0.3. Figure 2 pictorially demonstrates the lesion flagging threshold and also the difference between IoU and IoBB.

## Implementation details

Both detection models are initialized from pretrained weights. The backbone of VULD/ HPVD is initialized with ImageNet[40] pre-trained DenseNet121 weights. All other layers were randomly initialized. Specifically, the ImageNet pretrained models have red, green, and blue, three input channels. To migrate weights of the input layer for CT images, we reduce the three channels into one channel by taking the mean of the weights along the channel dimension. VULD/HPVD also comprise a feature pyramid network (FPN), a center point regression head, and a lesion size regression head, which are not in DenseNet121 and their network weights are assigned with the uniform initialization introduced by He *et al*[47]. HPVD updates VULD for hetero-model learning by replicating each of the BN layers in the first dense block into phase-specific BN layers and inserting the mean and variance operator to the end of the first dense block.
We implement the detection models in Pytorch on four NVIDIA Quadro RTX 6000 GPUs. To avoid using too much GPU memory, the input to the detection models are cropped CT images. Specifically, CT images are first resampled to the voxel spacing of 1mm by 1mm by 5mm with third order spline interpolation[48] and then normalized by their phase-specific HU mean and variance. In the training stage, a $$256 \times 256 \times 32$$ voxel sub-image is randomly cropped to feed into the network. Monitored on the validation set, the model was trained for 12,000 batches with a batch size of 4, sampling equally from all phase combinations. The Adam optimizer[49] was used with an initial learning rate of 0.0005 and then reduced by a factor of 10 after the first 6,000 batches. In the testing stage, a $$512 \times 512 \times 48$ window is the maximum size allowed by the GPU memory. If a testing image exceeds the maximum input size, a scanning window is applied across the depth dimension to process the whole testing image. Specifically, we crop along the depth dimension, creating sub-windows with an overlap of 16 voxels in the depth dimensions between consecutive sub-windows. The inference output in any overlapped regions is the mean of the two sub-windows.

## Reader Study

We recruited six clinical readers board certified in either China or Taiwan. These included two radiologists (Y.J.Z. and Y.T.H. with 6 and 12 years of experience, respectively), two HPB surgeons (Y.C.W. and C.H.L. with 9 and 12 years of experience, respectively), and two hepatologists (W.T. and C.W.P. with 9 and 5 years of experience, respectively). To reduce the burden of reading, we randomly selected 50 pathology-proven and 50 negative CT studies from the larger test set. Among the 50 pathology-proven studies, 45 were target studies containing one or more untreated HCC lesions and 5 were control studies with only TACE-treated HCC lesions. The demographic data of the reader study subset is shown in Supplementary Table S3. No identifying or additional information, outside of the CT scans themselves, were shared with the readers. We developed a plugin for the Medical Imaging Interaction Toolkit (MITK) viewer[50] that allowed readers to quickly navigate through a set of multi-phase CT studies and mark suspicious lesions with a RECIST marks, *i.e.*, to measure the size of the maximally suspicious finding with its long and short diameters on the key axial slice[51]. A 2D box can be generated from the resulting RECIST marks. Each reader was given the MITK software and the custom plugin, and we asked the readers to decide if any non-treated HCC lesions were present and to mark the maximally suspicious finding if so. For the same set of volumes, three rounds of reading was conducted, one for each contrast phase input setting. To keep no additional contrast information leaking into a subsequent round, the order of rounds was NC-only, NC + VP, and finally full DCE CT. 2D bboxes can be generated from the RECIST marks. However, due to data transfer errors, several studies had to re-annotated a second time by the readers. A localization was deemed correct if the mark passed the pointing game criterion[45] and there is a greater than 0.3 IoBB between the marked 2D bbox and the 2D bbox of any lesion projected from its ground truth 3D bbox onto the axial plane. Because readers were not asked to draw a 3D bbox, this localization criterion is more permissive (it is easier to obtain a 2D 0.3 IoBB than a 3D 0.3 IoBB) than the lesion flagging criterion used for the CADe models.

## Statistical analysis

We perform two primary types of analysis. The first is FROC analysis[14], common in CADe evaluations, which provides a measure of how well the CADe models can localize *all* lesion findings in the CT studies. To do this, we first apply NMS and then filter out all predicted TACE-treated localizations using the threshold classifier described above. What remains are a set of predicted lesion boxes along with a localization confidence for each. A threshold can be varied for the detection confidence, which filters out detected bboxes with a confidence lower than the threshold. A true positive bbox passes the lesion flagging criterion with any ground-truth lesion bbox. Otherwise a predicted bbox is a false positive (FP). The sensitivity and false-positives (FPs) per study are then measured across different confidence thresholds. Some liver lesion CADe works report sensitivities at FPs >= $1^{10,12}$. Instead, our FROC analysis reports the sensitivities at FPs ranging from 0 to 1 as FPs greater than 1 should not be considered clinically useful for HCC detection.

FROC analysis is informative, but it is less well known than ROC analysis and it does not provide a study-wise measure of whether studies with HCC lesions can be successfully flagged. Additionally, clinical readers were only asked to mark the maximally suspicious finding, if there was one. Thus, FROC metrics cannot be calculated for the reader study. Instead, we perform an LROC analysis[15], which only analyzes the predicted bbox with the highest confidence for each study. Instead of sensitivity, LROC analysis measures the TPLR, which is the proportion of target studies that had one of their lesions successfully localized by the maximally suspicious finding. The confidence threshold can be varied and the corresponding TPLR and specificities can be calculated. Note, unlike the sensitivity of ROC analysis, the maximum value of the TPLR may be less than 1.0 because the ground-truth lesions in some studies may not be localized by the maximally suspicious finding, i.e., the predicted bbox does not meet the lesion flagging criterion. These studies can never be localized no matter the confidence threshold. We use the AUC of the LROC as the main figure of merit. 95% confidence intervals and statistical significance tests on the improvements of the HPVD AUC over VULD were calculated using Wunderlich *et al.*'s non-parametric procedure. The Bonferroni correction was applied to correct for multiple comparisons of the latter[52].

# Competing Interests

The authors declare that there are no competing interests.

# Author Contribution

C.-T. C., A.P.H, L.L and C.-H.L. designed the experiments; W.T., Y-T. H., C.-T.C., and C.-H.L. acquired radiographics for use in the study and provided strategic support; J.C. , Y.T. and A. H. wrote code to achieve different tasks and carried out all experiments; J.C. implemented the annotation tools for data annotation; C.-T. C., Y.-C.W., C-W. P.and C.-H.L. provided labels for use in measuring algorithm performance; C.-T.C., J.C., L.L, A.H. and C-H. L. drafted the manuscript; T.-S. Y.. helped extensively with writing the manuscript; G.X., J.X., W.-C. L.and T.-

# Supplementary

## Supplementary method

The statistical analysis of the demographic distribution was computed using R version 3.6.3[1] with package "tableone."[2] We linked the Chang Gung Memorial Hospital pathology database and liver computed tomography (CT) study level data together. The patient's demographic data including age, gender, procedure of the specimen collection, hepatitis types, Ishak score, and cirrhosis were collected from the pathology report. The HCC lesion numbers, TACE lesion numbers, TACE present, HCC and TACE lesion co-occurrence, and total lesion number per study were calculated based on the annotation result per CT study. The manufacturer of the CT scanner, peak kilo voltage output of the X-Ray generator used (KVP), and the convolutional kernel applied in the target scan were collected through the DICOM attributes of each study. The continuous variables were compared with t-test or analysis of variance of more than two groups, and the categorical variables were compared with Fisher's exact test.

Table S1. Demographic distribution of the pathology-proven subsets of train, validation, and test data sets (note the training set is entirely pathology-proven, while the validation and test sets are augmented by negative non-pathology-proven controls). We use HCC to denote untreated HCC lesions, whereas TACE denotes TACE-treated HCC lesions. Pathology-proven cohorts can include studies with untreated HCC lesions (target) in addition to studies with only TACE-treated HCC lesions (controls). *p*-values test whether there are statistically significant differences between groups.

|  | Level | Train | Validation | Test | *p*-value |
|---|---|---|---|---|---|
| n |  | 771 | 99 | 185 |  |
| Total HCC lesion numbers |  | 851 | 107 | 175 |  |
| Total TACE lesion numbers |  | 73 | 32 | 50 |  |
| Age (sd) |  | 59.4 (12.3) | 59.4 (11.6) | 60.0 (10.8) | 0.831 |
| Gender (%) | Male | 619 (80.3) | 81 (81.8) | 139 (75.1) | 0.257 |
|  | Female | 152 (19.7) | 18 (18.2) | 46 (24.9) |  |
| Procedure, n (%) | Resection | 692 (89.8) | 82 (82.8) | 158 (85.4) | 0.048 |
|  | Transplant | 79 (10.2) | 17 (17.2) | 27 (14.6) |  |
| Hepatitis, n (%) | Hep B | 417 (54.1) | 53 (53.5) | 86 (46.5) | 0.367 |
|  | Hep B + Hep C | 36 ( 4.7) | 6 ( 6.1) | 9 ( 4.9) |  |
|  | Hep C | 147 (19.1) | 14 (14.1) | 46 (24.9) |  |
|  | Non-B Non-C | 56 (7.3) | 11 (11.1) | 17 (9.2) |  |
|  | Unknown | 115 (14.9) | 15 (15.2) | 27 (14.6) |  |
| Ishak score, mean (sd) |  | 4.17 (1.83) | 4.30 (1.67) | 4.26 (1.94) | 0.684 |
| Cirrhosis, n (%) | No | 378 (49.0) | 48 (48.5) | 81 (43.8) | 0.443 |
|  | Yes | 393 (51.0) | 51 (51.5) | 104 (56.2) |  |
| Lesion number per study, mean (sd) |  | 1.20 (0.58) | 1.40 (1.25) | 1.22 (0.76) | 0.023* |
| TACE present, n (%) | No | 713 (92.5) | 83 (83.8) | 157 (84.9) | <0.001* |
|  | Yes | 58 (7.5) | 16 (16.2) | 28 (15.1) |  |
| HCC number, n (%) | None | 10 (1.3) | 10 (10.1) | 21 (11.4) | <0.001* |
|  | Solitary | 688 (89.2) | 79 (79.8) | 155 (83.8) |  |
|  | Multiple | 73 (9.5) | 10 (10.1) | 9 (4.9) |  |
| HCC and TACE lesion co-occurrence, n (%) | No | 723 (93.8) | 93 (93.9) | 178 (96.2) | 0.447 |
|  | Yes | 48 (6.2) | 6 (6.1) | 7 (3.8) |  |
| Manufacturer, n (%) | GE | 120(15.6) | 20(20.2) | 26(14.1) | 0.309 |
|  | SIEMENS | 429(55.6) | 54(54.5) | 117(63.2) |  |
|  | TOSHIBA | 154(20.0) | 16(16.2) | 33(17.8) |  |
|  | OTHERS | 68(8.8) | 9(9.1) | 9(4.9) |  |
| KVP, n (%) | 120 | 712(92.3) | 93(93.9) | 173(93.5) | 0.976 |
|  | 140 | 56(7.3) | 6(6.1) | 12(6.5) |  |
|  | Others | 3(0.4) | 0(0.0) | 0(0.0) |  |
| Convolutional kernel, n(%) | GE Standard | 119(15.4) | 20(20.2) | 26(14.1) | 0.650 |
|  | Hitachi 84 | 29(3.8) | 3(3.0) | 4(2.2) |  |
|  | Hitachi 85 | 38(4.9) | 6(6.1) | 4(2.2) |  |
|  | SIEMENS B30f | 107(13.9) | 7(7.1) | 29(15.7) |  |
|  | SIEMENS B31f | 308(39.9) | 44(44.4) | 84(45.4) |  |
|  | TOSHIBA FC04 | 36(4.7) | 4(4.0) | 6(3.2) |  |

| | | | |
|---|---|---|---|
| TOSHIBA FC08 | 26(3.4) | 2(2.0) | 8(4.3) |
| TOSHIBA FC18 | 68(8.8) | 7(7.1) | 16(8.6) |
| Others | 40(5.2) | 6(6.1) | 8(4.3) |

\* statistically significant.. *sd* standard deviation, *Hep* hepatitis, *HCC* hepatocellular carcinoma, *TACE* transarterial chemoembolization, *KVP* peak kilo voltage output of the X-Ray generator used.

Table S2. The detailed demographic information of the test set, including non-pathology-proven negative controls.

|  | Level | Target | Control | p-value |
|---|---|---|---|---|
| n |  | 164 | 206 |  |
| Age, mean years (sd) |  | 59.82(11.15) | 53.98(13.68) | <0.001* |
| Gender, n (%) | Male | 125(76.2) | 131(63.6) | 0.009* |
|  | Female | 39(23.8) | 75(36.4) |  |
| TACE present, n (%) | No | 157(95.7) | 185(89.8) | 0.046 |
|  | Yes | 7(4.3) | 21(10.2) |  |
| Manufacturer, n (%) | GE | 24(14.6) | 43(20.9) | <0.001* |
|  | SIEMENS | 102(62.2) | 107(51.9) |  |
|  | TOSHIBA | 29(17.7) | 56(27.2) |  |
|  | OTHERS | 9(5.5) | 0(0.0) |  |
| KVP, n (%) | 120 | 153(93.3) | 203(98.5) | 0.002* |
|  | 140 | 11(6.7) | 1(0.5) |  |
|  | Others | 0(0.0) | 2(1.0) |  |
| Convolutional kernel, n(%) | GE Standard | 24(14.6) | 42(20.4) | <0.001* |
|  | Hitachi 84 | 4(2.4) | 0(0.0) |  |
|  | Hitachi 85 | 4(2.4) | 0(0.0) |  |
|  | SIEMENS B30f | 26(15.9) | 14(6.8) |  |
|  | SIEMENS B31f | 72(43.9) | 88(42.7) |  |
|  | TOSHIBA FC04 | 6(3.7) | 2(1.0) |  |
|  | TOSHIBA FC08 | 8(4.9) | 13(6.3) |  |
|  | TOSHIBA FC18 | 13(7.9) | 39(18.9) |  |
|  | Others | 7(4.3) | 8(3.9) |  |

* statistically significant. *sd* standard deviation, *Hep* hepatitis, *HCC* hepatocellular carcinoma, *TACE* transarterial chemoembolization, *KVP* peak kilo voltage output of the X-Ray generator used.

Table S3. The detailed demographic information of the reader study test set, including non-pathology-proven negative controls.

|  | level | Target | Control | p-value |
|---|---|---|---|---|
| n |  | 45 | 55 |  |
| Age, mean years (sd) |  | 57.72(9.65) | 52.80(12.83) | 0.036* |
| Gender, n (%) | Male | 33(73.3) | 34(61.8) | 0.286 |
|  | Female | 12(26.7) | 21(38.2) |  |
| TACE present, n (%) | No | 42(93.3) | 50(90.9) | 0.727 |
|  | Yes | 3(6.7) | 5(9.1) |  |
| Manufacturer, n (%) | GE | 3(6.7) | 14(25.5) | 0.004* |
|  | SIEMENS | 30(66.7) | 22(40.0) |  |
|  | TOSHIBA | 10(22.2) | 19(34.5) |  |
|  | OTHERS | 2(4.4) | 0(0.0) |  |
| KVP, n (%) | 120 | 40(88.9) | 54(98.2) | 0.088 |
|  | 140 | 5(11.1) | 1(1.8) |  |
| Convolutional kernel, n(%) | GE Standard | 3(6.7) | 14(25.5) | 0.002* |
|  | Hitachi 84 | 1(2.2) | 0(0.0) |  |
|  | SIEMENS B30f | 11(24.4) | 5(9.1) |  |
|  | SIEMENS B31f | 18(40.0) | 17(30.9) |  |
|  | TOSHIBA FC04 | 2(4.4) | 0(0.0) |  |
|  | TOSHIBA FC08 | 5(11.1) | 6(10.9) |  |
|  | TOSHIBA FC18 | 3(6.7) | 13(23.6) |  |
|  | Others | 2(4.4) | 0(0.0) |  |

* statistically significant. *sd* standard deviation *Hep* hepatitis, *HCC* hepatocellular carcinoma, *TACE* transarterial chemoembolization, *KVP* peak kilo voltage output of the X-Ray generator used.

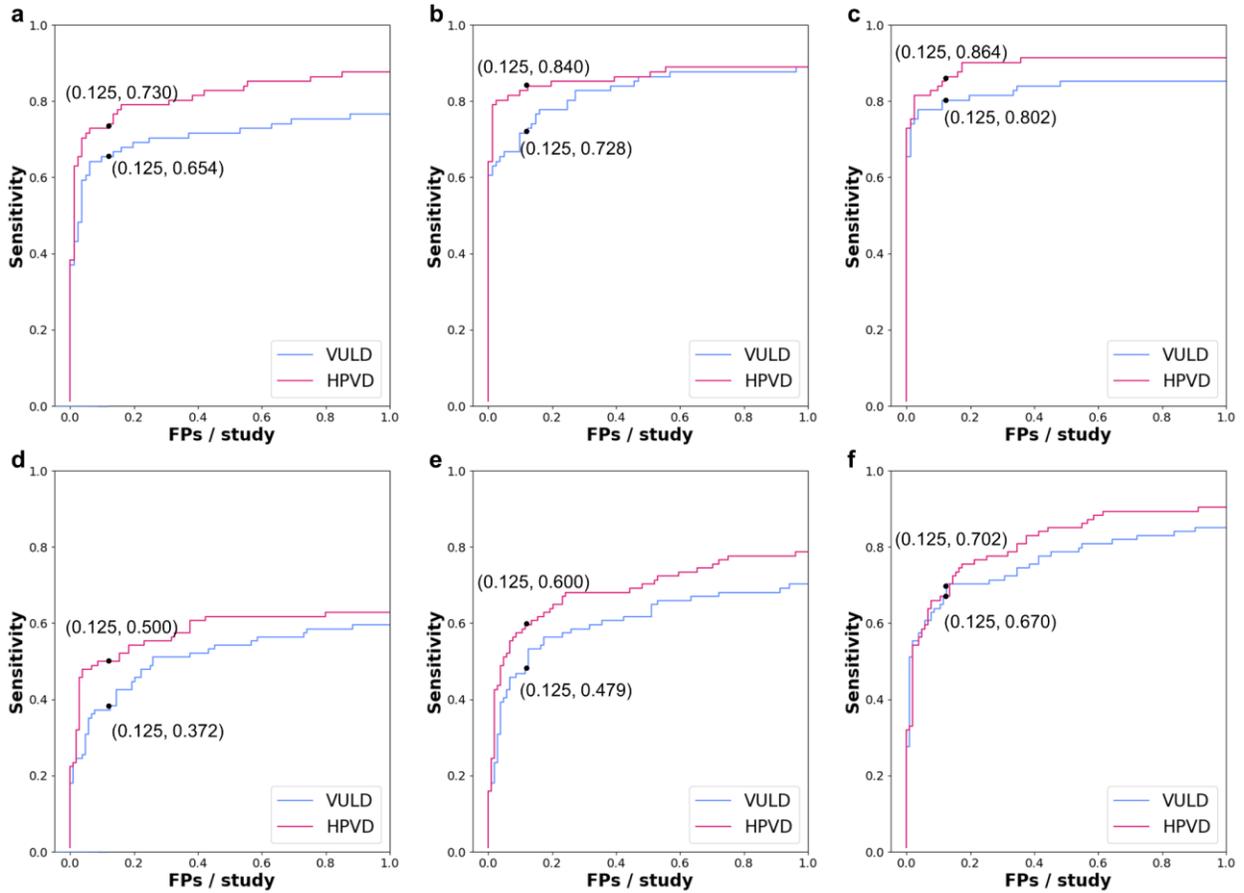

Figure S1. **FROC curves of detection performance of the VULD and HPVD models.** Because only liver fibrosis severity was not known for non-pathology-proven controls, results are only presented on the pathology-proven test subset (n=185). Row 1 depicts performance of patients (n=81) whose histopathology-derived Ishak scores range from 0 to 4, *i.e.*, no cirrhosis. Plots **a**, **b**, and **c** depict detection performance on NC-only, NC + VP, and full DCE CT studies, respectively. Row 2 depicts performance of cirrhotic patients (n=104), *i.e.*, those with histopathology-derived Ishak scores to be either 5 or 6. Plots **d**, **e**, and **f** depict detection performance on NC-only, NC + VP, and full DCE CT studies, respectively.

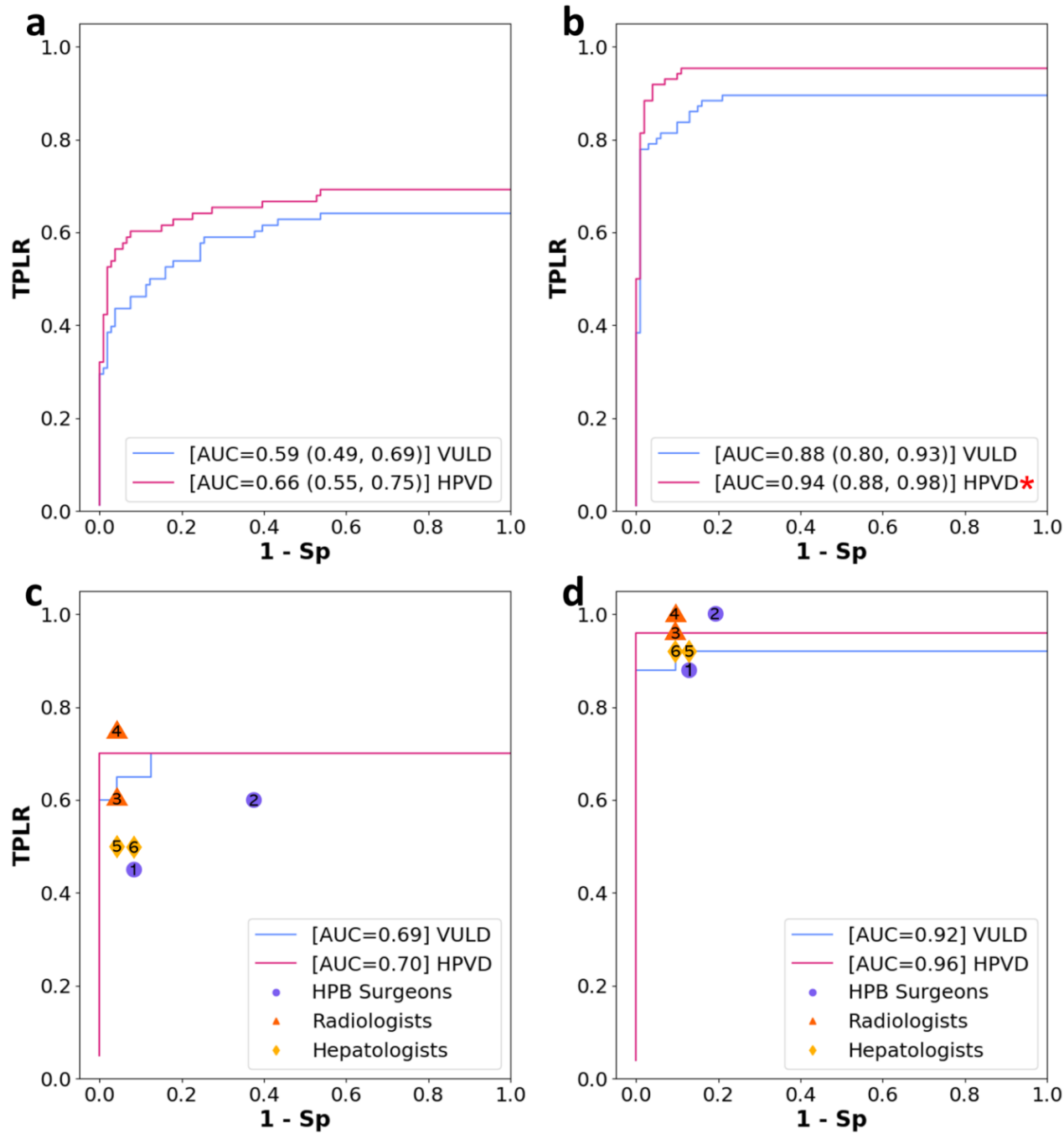

Figure S2. **LROC curves using NC + VP studies stratified by the size of the largest lesion.** **a** and **b** depict results on the full test set for studies with small (n=184) and large (n=186) lesions, respectively. AUCs are reported with 95% confidence intervals. The red asterisk mark in **b** indicates the differences between the VULD and HPVD AUCs are statistically significant. **c** and **d** depict results on the reader study subset for studies with small (n=44) and large (n=56) lesions, respectively, with reader performance also indicated.

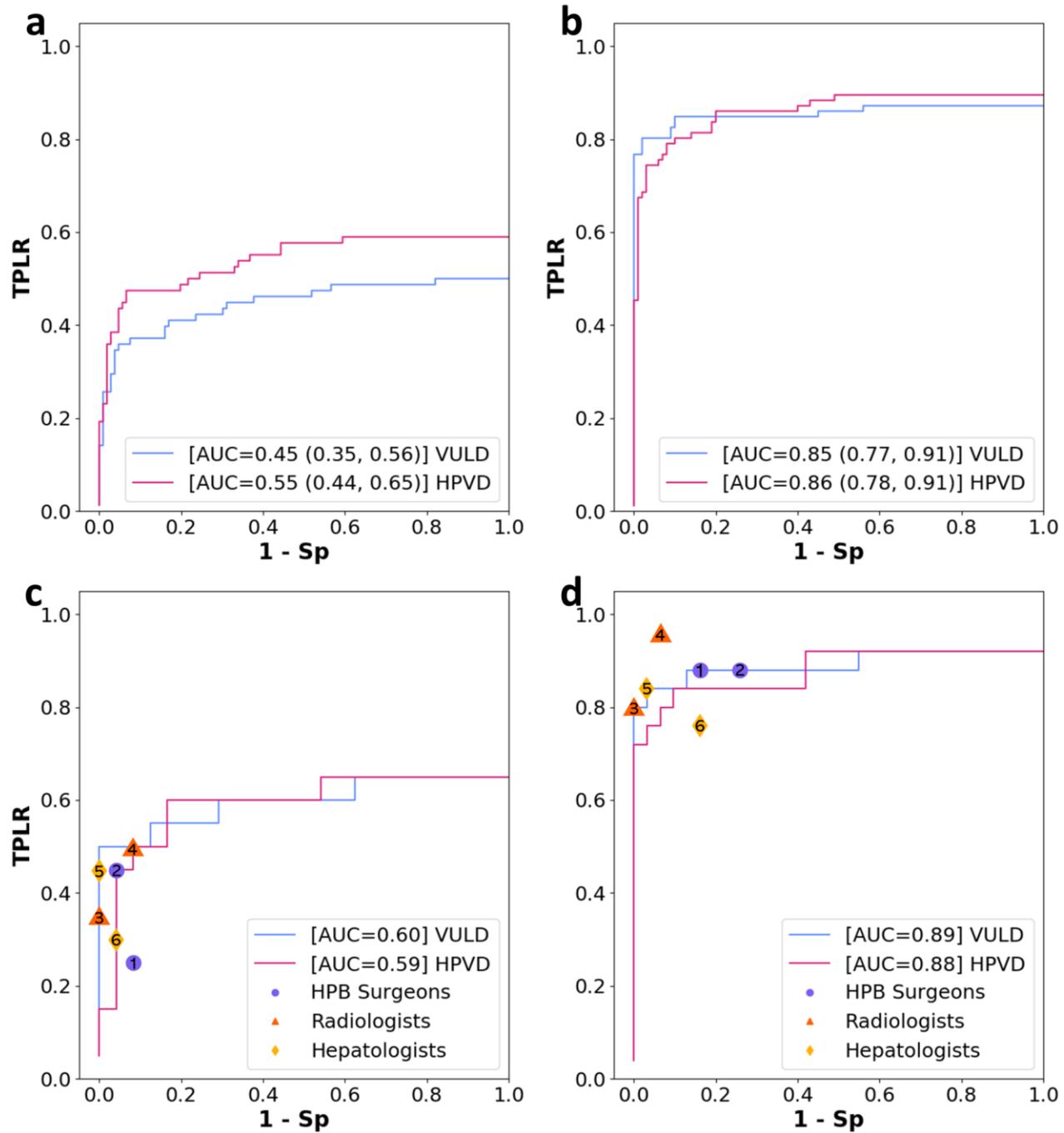

Figure S3. **LROC curves using NC-only studies stratified by the size of the largest lesion. a** and **b** depict results on the full test set for studies with small (n=184) and large (n=186) lesions, respectively. AUCs are reported with 95% confidence intervals. **c** and **d** depict results on the reader study subset for studies with small (n=44) and large (n=56) lesions, respectively, with reader performance also indicated.